\pdfoutput=1

\documentclass[11pt]{article}

\usepackage{ACL2023}

\usepackage{times}
\usepackage{latexsym}
\usepackage{graphicx}

\usepackage[T1]{fontenc}

\usepackage[utf8]{inputenc}

\usepackage{microtype}

\usepackage{inconsolata}

\title{Overview  of the VLSP 2022 - Abmusu Shared Task: A Data Challenge for Vietnamese Abstractive Multi-document Summarization}

\author{
    \textbf{Mai-Vu Tran, Hoang-Quynh Le\thanks{$^*$Corresponding author}, Duy-Cat Can, Quoc-An Nguyen} \\ 
    VNU University of Engineering and Technology, Hanoi, Vietnam.\\
    {\tt\{vutm, lhquynh, catcd, annq\}@vnu.edu.vn}\\
}

\begin{document}
\maketitle
\begin{abstract}
This paper reports the overview of the VLSP 2022 - Vietnamese abstractive multi-document summarization (Abmusu) shared task for Vietnamese News. This task is hosted at the 9$^{th}$ annual workshop on Vietnamese Language and Speech Processing (VLSP 2022). 
The goal of Abmusu shared task is to develop summarization systems that could create abstractive summaries automatically for a set of documents on a topic. The model input is multiple news documents on the same topic, and the corresponding output is a related abstractive summary. In the scope of Abmusu shared task, we only focus on Vietnamese news summarization and build a human-annotated dataset of 1,839 documents in 600 clusters, collected from Vietnamese news in 8 categories. 
Participated models are evaluated and ranked in terms of \texttt{ROUGE2-F1} score, the most typical evaluation metric for document summarization problem. 
\end{abstract}

\begin{figure*}[!ht]
	\centering
	\includegraphics[width=0.9\linewidth]{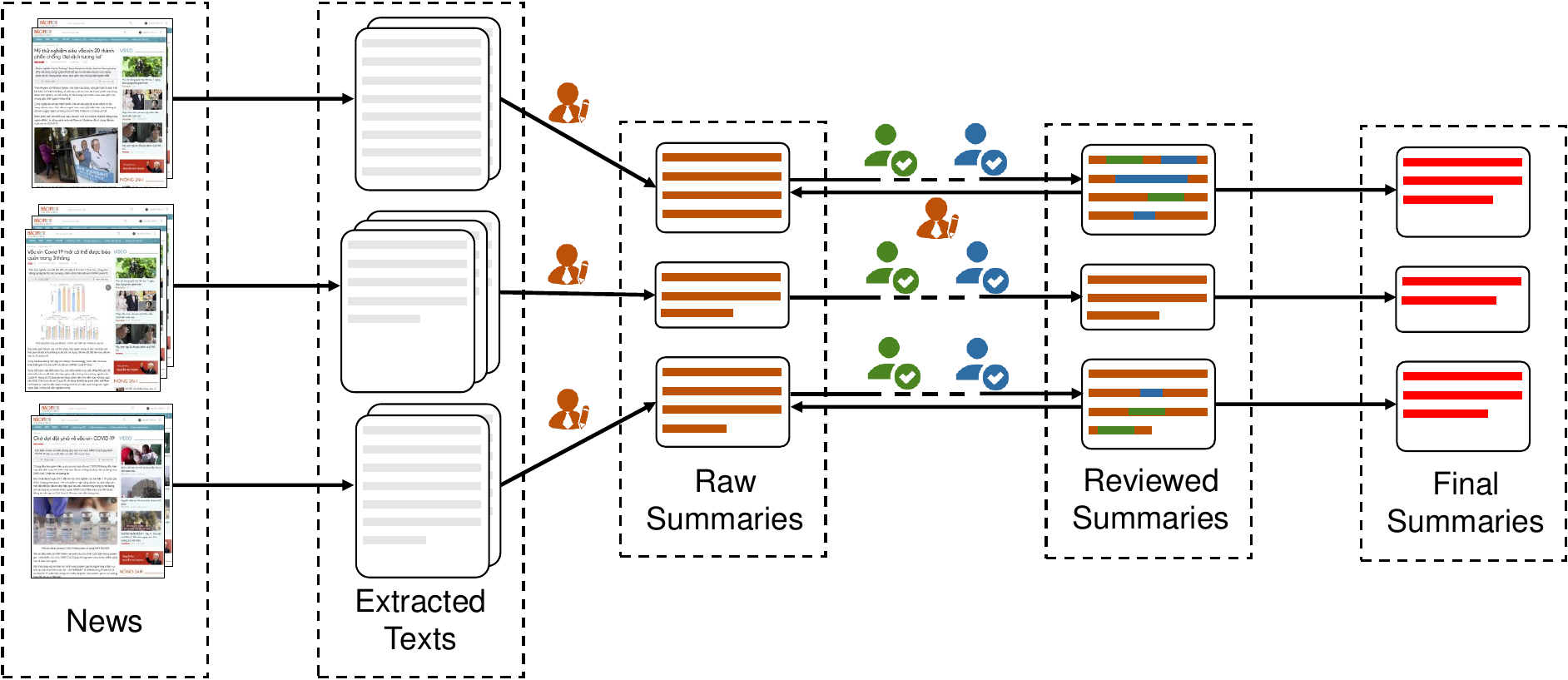}
	\caption{The annotation process.}
	\label{fig:datapreparation}
\end{figure*}

\section{Introduction}

In the era of information explosion, mining data effectively has huge potential but is a difficult problem which takes time, money and labour effort. Multi-document summarization is a natural language processing task that is useful for solving this problem. Receiving the set of documents as input, the summarization system aims to select or generate important information to create a brief summary for these documents \cite{jevzek2008automatic}. It is a complex problem that has gained attention from the research community. 
Several past challenges and shared tasks have focused on summarization. One of the earliest summarization shared tasks is the series of document understanding conference (DUC) challenges\footnote{\url{http://www-nlpir.nist.gov/projects/duc}. DUC summarization challenges are organized 7 times from 2000 to 2007.}, the Text Analysis Conference (TAC) summarization shared tasks\footnote{\url{http://tac.nist.gov/tracks/}. TAC summarization shared tasks are organized 5 times on summarization news and biomedical text from 2008 to 2014.}
In recent years, some summarization shared tasks have been launched to support research and development in this field for English, such as DocEng~2019 \cite{lins2019doceng} and BioNLP-MEDIQA~2021 \cite{abacha2021overview}, ect.

Based on output characteristics, there are two major approaches for automatic summarization, i.e, extractive and abstractive summarization. Extractive summarization tends to select the most crucial sentences (sections) from the documents while abstractive summarization tries to rewrite a new summary based on the original important information \cite{allahyari2017text}. From the early 1950s, various methods have been proposed for extractive summarization ranging from frequency-based methods \cite{khan2019extractive} to machine learning-based methods \cite{gambhir2017recent}. The extractive methods are fast and simple but the summaries are far from the manual-created summary, which can be remedied with the abstractive approach \cite{el2021automatic}. In the multi-document problem, extractive approaches show significant disadvantages in arranging and combining information from several documents. In recent years, sequence-to-sequence learning (seq2seq) makes abstractive summarization possible \cite{hou2017abstractive}. A set of models based on encoder-decoder such as PEGASUS \cite{zhang2020pegasus}, BART \cite{lewis2020bart}, T5 \cite{raffel2020exploring} achieves potential results for abstractive multi-document summarization. 
Studies on this problem for Vietnamese text are still in the early stages with a few initial achievements, especially in extractive approaches. In recent years, there has been a growing interest to develop automatic abstractive summarization systems. Despite these attempts, the lack of a comprehensive benchmarking dataset has limited the comparison of different techniques for Vietnamese. VLSP 2022 - Abmusu shared task is set up to provide an opportunity for researchers to propose, assess and advance their research, further, promote the development of research on abstractive multi-document summarization for Vietnamese text.

The remainder of the paper is organized as follows: 
Section 2 gives a detailed description of the Abmusu shared task and the task data.
The next section describes the data construction, annotation methodologies and data collection.
Section 3 describes the competition, baselines, approaches and respective results. 
Finally, Section 4 concludes the paper.

\section{Task Description}
VLSP 2022 Abmusu shared task addressed an abstractive multi-document summarization task.
The goal of Abmusu shared task is to develop summarization systems that could create abstractive summaries automatically for a set of documents on a topic. The model input is multiple news documents on the same topic, and the corresponding output is a related abstractive summary. In the scope of Abmusu shared task, we only focus on Vietnamese news.
For multi-document summarization purposes, Abmusu task is aimed at summarizing multiple input documents that contain a piece of information related to the same topic, we call them \textit{`document clusters'}. Each cluster has $3-5$ documents that illustrate the same topic and the goal of this shared task is to build models to create an abstractive summary per cluster automatically. 

\section{Task Data}
\subsection{Data Preparation}
The data is automatically collected and filtered from Vietnamese electronic news on $8$ categories, including the economy, society, culture, science and technology, etc. It is divided into training, validation and test datasets. The datasets contain several document clusters. Each cluster has $3-5$ documents that illustrate the same topic. On training and validation datasets, a manual-created reference abstractive summary is provided per cluster.
The test set is formatted similarly to the training and validation sets, but without an abstractive summary. 

The data preparation process is described in Figure~\ref{fig:datapreparation}. We used INCEpTION\footnote{\url{http://https://inception-project.github.io//}} \cite{jan2018inception} as the annotation tool. It is a semantic annotation platform offering intelligent assistance and knowledge management.
There are $10$ human annotators and $2$ experts who participated in the annotation process, the annotation guideline with full definition and illustrative examples was provided. 
We used an $8-$step process to make summarization data, each data sample needs the involvement of at least $1$ annotator and $1$ reviewer:
\begin{itemize}
    \item Crawl data from news websites by categories.
    \item Group documents into clusters by the highlighted hashtag, category, posted time, and similarity.
    \item Remove duplicate or highly similar documents.
    \item Remove clusters with too few articles, and review to select clusters/documents manually.
    \item Choose $200$ more clusters randomly to ensure the distribution for difficult test-cases.
    \item Create the summary manually by the annotators.
    \item Re-check the quality of the summary (by the reviewers) to ensure the quality and length. Unqualified data is relabeled by another annotator.
    \item Refine all data by expert reviewers.
\end{itemize}

\begin{figure*}[!ht]
	\centering
	\includegraphics[width=\linewidth]{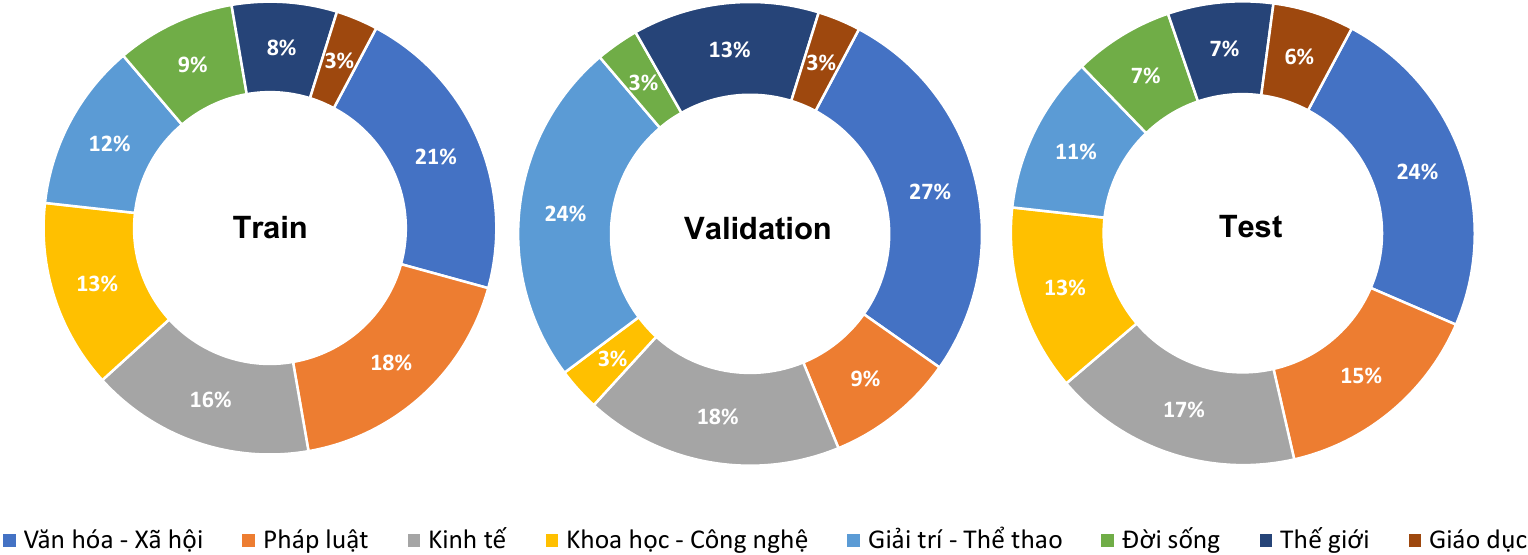}
	\caption{The data statistics by categories.}
	\label{fig:datapiechart}
\end{figure*}

As a result, we prepared a total of $1,839$ documents in $600$ clusters: $621$ documents ($200$ clusters) for the training set, $304$ documents ($100$ clusters) for the validation set and $914$ documents ($300$ clusters) in the test set.
Figure~\ref{fig:datapiechart} show the distribution of categories in the training/validation set and the test set.
Table~\ref{tab:data1} and Table~\ref{tab:data2} describe the statistics of the Abmusu dataset in detail at the token- and the sentence level. 
The compression ratio of Abmusu dataset is $\sim9\%$, the manually created summaries often contain $4-6$ sentences.

\begin{table}[h]
\centering
\scalebox{0.75}{
\begin{tabular}{|lccc|}
\hline
\multicolumn{1}{|l|}{\textbf{Aspects}}                & \multicolumn{1}{c|}{\textbf{Training}} & \multicolumn{1}{c|}{\textbf{Validation}} & \textbf{Test}    \\ \hline
\multicolumn{4}{|l|}{\textbf{Average}}                                                                                            \\ \hline
\multicolumn{1}{|l|}{Documents per Cluster}  & \multicolumn{1}{c|}{3.11}     & \multicolumn{1}{c|}{3.04}       & 3.05    \\ \hline
\multicolumn{1}{|l|}{Tokens per Cluster}     & \multicolumn{1}{c|}{1924.75}  & \multicolumn{1}{c|}{1815.41}    & 1762.40 \\ \hline
\multicolumn{1}{|l|}{Tokens per Raw text}    & \multicolumn{1}{c|}{619.88}   & \multicolumn{1}{c|}{597.17}     & 578.46  \\ \hline
\multicolumn{1}{|l|}{Tokens per Anchor text} & \multicolumn{1}{c|}{41.65}    & \multicolumn{1}{c|}{35.58}      & 40.33   \\ \hline
\multicolumn{1}{|l|}{Tokens per Summary}     & \multicolumn{1}{c|}{168.48}   & \multicolumn{1}{c|}{167.68}     & 153.05  \\ \hline
\multicolumn{4}{|l|}{\textbf{Compression ratio}}                                                                                  \\ \hline
\multicolumn{1}{|l|}{Multi-document Summary} & \multicolumn{1}{c|}{0.09}     & \multicolumn{1}{c|}{0.09}       & 0.09    \\ \hline
\end{tabular}
}
\caption{Average statistics and compression ratio at token-level}
\label{tab:data1}
\end{table}

\begin{table}[h]
\centering
\scalebox{0.75}{
\begin{tabular}{|lccc|}
\hline
\multicolumn{1}{|l|}{\textbf{Aspects}}                   & \multicolumn{1}{c|}{\textbf{Training}} & \multicolumn{1}{c|}{\textbf{Validation}} & \textbf{Test}  \\ \hline
\multicolumn{4}{|l|}{\textbf{Average}}                                                                                             \\ \hline
\multicolumn{1}{|l|}{Sentences per Cluster}     & \multicolumn{1}{c|}{66.93}    & \multicolumn{1}{c|}{60.69}      & 61.07 \\ \hline
\multicolumn{1}{|l|}{Sentences per Raw text}    & \multicolumn{1}{c|}{21.56}    & \multicolumn{1}{c|}{19.96}      & 20.04 \\ \hline
\multicolumn{1}{|l|}{Sentences per Anchor text} & \multicolumn{1}{c|}{1.72}     & \multicolumn{1}{c|}{1.27}       & 1.57  \\ \hline
\multicolumn{1}{|l|}{Sentences per Summary}     & \multicolumn{1}{c|}{4.82}     & \multicolumn{1}{c|}{4.94}       & 4.93  \\ \hline
\multicolumn{4}{|l|}{\textbf{Compression ratio}}                                                                                   \\ \hline
\multicolumn{1}{|l|}{Multi-document Summary}    & \multicolumn{1}{c|}{0.07}     & \multicolumn{1}{c|}{0.08}       & 0.08  \\ \hline
\end{tabular}
}
\caption{Average statistics and compression ratio at sentence-level}
\label{tab:data2}
\end{table}

\section{Challenge Results}
\subsection{Data Format and Submission}
Each data example includes the title, anchor text and body text of all single documents in a cluster. Each cluster also has a category tag and a manually created summary. 
The provided test set for the participated team is formatted similarly to the training and validation data, but without the manually created summary. 
The evaluation was performed on the AIhub\footnote{\url{http://aihub.ml/}} platform for $7$ days. Test data was divided into two parts: Public Test and Private Test, each containing $50\%$ of the test data. The Private Test was opened $4$ days after the Public Test. 
Each team is allowed to submit a maximum of $35$ submissions to the Public test ($5$ per day) and $5$ submissions to the Private Test (not limited per day). 

\subsection{Evaluation Metrics}
The official evaluation measures are the \texttt{ROUGE-2} scores and \texttt{ROUGE-2 F1 (R2-F1)} is the main score for ranking. \texttt{ROUGE-2 Recall (R2-R)}, \texttt{Precision (R2-P)} and \texttt{R2-F1} between predicted summary and reference summary are calculated as \cite{lin2004rouge}:

\begin{equation}
\texttt{R2-P} = \frac{|\texttt{Matched~n-grams}|}{|\texttt{Predicted~summary~n-grams}|}
\end{equation}

\begin{equation}
\texttt{R2-R} = \frac{|\texttt{Matched~n-grams}|}{|\texttt{Reference ~summary~n-grams}|}
\end{equation}

\begin{equation}
\texttt{R2-F1} = \frac{\texttt{2}\times\texttt{R2-P}\times\texttt{R2-R}}{\texttt{R2-P}+\texttt{R2-R}}
\end{equation}

\subsection{Baselines}
The committee provided $4$ baselines as the shared task benchmark, includes:
\begin{itemize}
    \item Ad-hoc rule-based baseline: The summary is the concatenation of the first and the last sentences of all component documents in each cluster.
    \item Anchor text-based baseline: The summary is the concatenation of the anchor text of all component documents in each cluster.
    \item Extractive baseline: The summary is generated by the extractive summarization model using Lexrank \cite{erkan2004lexrank} and MMR \cite{goldstein1998summarization}.
    \item Abstractive baseline: The summary is generated by the abstractive summarization model ViT5 \cite{phan2022vit5}.
\end{itemize}

\subsection{Participants}
There are $46$ registered teams from research groups in domestic and international Universities (VNU-HUS, VNU-UET, HUST, PTIT, etc.) and industries (Viettel, VinGroup, CMC, TopCV, VCCorp, etc). 
In which, 28 teams submitted the data agreement, and 16 teams participated officially by submitting at least 1 run on the evaluation platform.
Participant teams can use all possible tools and resources to build models.
Participated teams made a total of~287 submissions.
Post-challenge panels\footnote{\url{http://aihub.ml/competitions/341}} are now opened on AIHUB for supporting research improvements.

\subsection{Results}
An interesting observation is that the rule-based baseline achieved surprisingly high results (ranked~6). This result can be explained because most news are written in an explanatory or inductive style, so the first and last sentences often contain important information. The extractive baseline result (ranked~5) was much better than the anchor text baseline result (ranked~18), contrary to the assumption that the anchor text can be considered as a simple summary of the news text.
In the abstractive baseline model, we only put raw data through the ViT5 model without any parameter tuning, so it is reasonable when its result was low (ranked~19). 

The proposed models followed two main approaches: abstractive summarization and hybrid approach. Participated teams used a variety of techniques, including similarity scoring (TF-IDF, Cosine, etc,), graph-based methods (i.e., Lexrank \cite{erkan2004lexrank}, Textrank \cite{mihalcea2004textrank}, Pagerank \cite{brin1998anatomy},~etc.), sentence classification (Long short-term memory \cite{hochreiter1997long}, BERT \cite{kenton2019bert}, etc.) and text correlation.
The results of the private test were considered as the official results to rank the team in Abmusu shared task. The results on \texttt{ROUGE-2} of the top 5 teams and 4 baselines are shown on Table~\ref{tab:result1} (See Appendix~\ref{sec:appendix} for the full results.).
All $16$ teams achieved performance higher than the anchor text baseline and abstractive baseline.
There were $5$ teams that achieved a higher F-score than our extractive and rule-based baselines.
The best \texttt{ROUGE-2~F} obtained was $0.3035$, the corresponding \texttt{ROUGE-2~P} and \texttt{ROUGE-2~R} are $0.3035$ and $0.2298$ respectively. 

\begin{table}[h]
\centering
\resizebox{\linewidth}{!}{%
\setlength{\tabcolsep}{10pt}
\begin{tabular}{|c|l|c|c|c|}
\hline
\textbf{Rank} & \multicolumn{1}{c|}{\textbf{Team}} & \textbf{\texttt{R2-F1}} & \textbf{\texttt{R2-P}} & \textbf{\texttt{R2-R}} \\ \hline
1 & LBMT & \textbf{\begin{tabular}[c]{@{}c@{}}0.3035\\ (1)\end{tabular}} & \begin{tabular}[c]{@{}c@{}}0.2298\\ (11)\end{tabular} & \textbf{\begin{tabular}[c]{@{}c@{}}0.4969\\ (1)\end{tabular}} \\ \hline
2 & The coach & \begin{tabular}[c]{@{}c@{}}0.2937\\ (2)\end{tabular} & \begin{tabular}[c]{@{}c@{}}0.2284\\ (12)\end{tabular} & \begin{tabular}[c]{@{}c@{}}0.4463\\ (2)\end{tabular} \\ \hline
3 & CIST AI & \begin{tabular}[c]{@{}c@{}}0.2805\\ (3)\end{tabular} & \begin{tabular}[c]{@{}c@{}}0.2629\\ (6)\end{tabular} & \begin{tabular}[c]{@{}c@{}}0.3192\\ (6)\end{tabular} \\ \hline
4 & TheFinalYear & \begin{tabular}[c]{@{}c@{}}0.2785\\ (4)\end{tabular} & \begin{tabular}[c]{@{}c@{}}0.2272\\ (13)\end{tabular} & \begin{tabular}[c]{@{}c@{}}0.4040\\ (4)\end{tabular} \\ \hline
5 & NLP HUST & \begin{tabular}[c]{@{}c@{}}0.2689\\ (5)\end{tabular} & \begin{tabular}[c]{@{}c@{}}0.2773\\ (4)\end{tabular} & \begin{tabular}[c]{@{}c@{}}0.2829\\ (12)\end{tabular} \\ \hline
\textit{6} & \textit{Extractive baseline} & \textit{\begin{tabular}[c]{@{}c@{}}0.2625\\ (6)\end{tabular}} & \textit{\begin{tabular}[c]{@{}c@{}}0.2464\\ (7)\end{tabular}} & \textit{\begin{tabular}[c]{@{}c@{}}0.3174\\ (8)\end{tabular}} \\ \hline
\textit{7} & \textit{Rule based baseline} & \textit{\begin{tabular}[c]{@{}c@{}}0.2611\\ (7)\end{tabular}} & \textit{\begin{tabular}[c]{@{}c@{}}0.2634\\ (5)\end{tabular}} & \textit{\begin{tabular}[c]{@{}c@{}}0.2947\\ (10)\end{tabular}} \\ \hline
\textit{19} & \textit{Anchor baseline} & \textit{\begin{tabular}[c]{@{}c@{}}0.1886\\ (18)\end{tabular}} & \textit{\begin{tabular}[c]{@{}c@{}}0.2306\\ (10)\end{tabular}} & \textit{\begin{tabular}[c]{@{}c@{}}0.1734\\ (19)\end{tabular}} \\ \hline
\textit{20} & \textit{Abstractive baseline} & \textit{\begin{tabular}[c]{@{}c@{}}0.1497\\ (19)\end{tabular}} & \textit{\textbf{\begin{tabular}[c]{@{}c@{}}0.3061\\ (1)\end{tabular}}} & \textit{\begin{tabular}[c]{@{}c@{}}0.1025\\ (20)\end{tabular}} \\ \hline
\end{tabular}

}

\caption{The official top 5 results on the Private Test. The number highlighted in bold is the highest result in each column. The number in the bracket () is the corresponding rank of a score. Baseline results are shown in italic.}
\label{tab:result1}
\end{table}

\section{Conclusions}
The VLSP 2022 - Abmusu shared task was designed to promote the development of research for the problem of abstractive multi-document summarization problem. We tend to compare different summarization approaches and provide a standard test-bed for future research.
The Abmusu dataset is constructed carefully, it is expected to make significant contributions to the other related works.  
Abmusu attracted the attention of the research community, participated teams came up with many different approaches and used a variety of advanced technologies and resources. We archived some exciting and potential results, which are useful benchmarks for future research.
Finally, we happily conclude that the VLSP 2022 - Abmusu shared task was run successfully and is expected to contribute significantly to Vietnamese text mining and natural language processing communities.

\section*{Acknowledgments}
The data was supported by the Project ``Research and Development of Vietnamese Multi-document Summarization Based on Advanced Language Models'' of Vietnam National University, Hanoi (Code: QG.22.61). 
The shared task committee would like to gratitude Dagoras Technology and Communications JSC. for their technical and financial support. 
We also thank all members of the Data Science and Knowledge Technology Laboratory, FIT, UET, VNU because of their continuous support and encouragement.

\bibliography{anthology,custom}
\bibliographystyle{acl_natbib}

\onecolumn
\appendix

\section{Appendix: The official results on the Private test}
\label{sec:appendix}

\begin{table*}[h]
\centering
\resizebox{\linewidth}{!}{%
\setlength{\tabcolsep}{10pt}

\begin{tabular}{|c|l|c|c|c|c|c|c|c|c|c|}
\hline
\textbf{Rank} & \multicolumn{1}{c|}{\textbf{Team}} & \textbf{\texttt{R2-F1}}                                                 & \textbf{\texttt{R2-P}}                                                          & \textbf{\texttt{R2-R}}                                                  & \textbf{\texttt{R1-F1}}                                                 & \textbf{\texttt{R1-P}}                                                          & \textbf{\texttt{R1-R}}                                                  & \textbf{\texttt{RL-F1}}                                                 & \textbf{\texttt{RL-P}}                                                 & \textbf{\texttt{RL-R}}                                                  \\ \hline
1             & LBMT                               & \textbf{\begin{tabular}[c]{@{}c@{}}0.3035\\ (1)\end{tabular}}  & \begin{tabular}[c]{@{}c@{}}0.2298\\ (11)\end{tabular}                  & \textbf{\begin{tabular}[c]{@{}c@{}}0.4969\\ (1)\end{tabular}}  & \textbf{\begin{tabular}[c]{@{}c@{}}0.5067\\ (1)\end{tabular}}  & \begin{tabular}[c]{@{}c@{}}0.4076\\ (16)\end{tabular}                  & \textbf{\begin{tabular}[c]{@{}c@{}}0.7147\\ (1)\end{tabular}}  & \textbf{\begin{tabular}[c]{@{}c@{}}0.4809\\ (1)\end{tabular}}  & \begin{tabular}[c]{@{}c@{}}0.3868\\ (15)\end{tabular}         & \textbf{\begin{tabular}[c]{@{}c@{}}0.6780\\ (1)\end{tabular}}  \\ \hline
2             & The coach                          & \begin{tabular}[c]{@{}c@{}}0.2937\\ (2)\end{tabular}           & \begin{tabular}[c]{@{}c@{}}0.2284\\ (12)\end{tabular}                  & \begin{tabular}[c]{@{}c@{}}0.4463\\ (2)\end{tabular}           & \begin{tabular}[c]{@{}c@{}}0.4962\\ (2)\end{tabular}           & \begin{tabular}[c]{@{}c@{}}0.4072\\ (17)\end{tabular}                  & \begin{tabular}[c]{@{}c@{}}0.6676\\ (4)\end{tabular}           & \begin{tabular}[c]{@{}c@{}}0.4701\\ (2)\end{tabular}           & \begin{tabular}[c]{@{}c@{}}0.3857\\ (16)\end{tabular}         & \begin{tabular}[c]{@{}c@{}}0.6326\\ (4)\end{tabular}           \\ \hline
3             & CIST AI                            & \begin{tabular}[c]{@{}c@{}}0.2805\\ (3)\end{tabular}           & \begin{tabular}[c]{@{}c@{}}0.2629\\ (6)\end{tabular}                   & \begin{tabular}[c]{@{}c@{}}0.3192\\ (6)\end{tabular}           & \begin{tabular}[c]{@{}c@{}}0.4876\\ (4)\end{tabular}           & \begin{tabular}[c]{@{}c@{}}0.4635\\ (6)\end{tabular}                   & \begin{tabular}[c]{@{}c@{}}0.5352\\ (9)\end{tabular}           & \begin{tabular}[c]{@{}c@{}}0.4541 \\ (4)\end{tabular}          & \begin{tabular}[c]{@{}c@{}}0.4314\\ (6)\end{tabular}          & \begin{tabular}[c]{@{}c@{}}0.4988\\ (7)\end{tabular}           \\ \hline
4             & TheFinalYear                       & \begin{tabular}[c]{@{}c@{}}0.2785\\ (4)\end{tabular}           & \begin{tabular}[c]{@{}c@{}}0.2272\\ (13)\end{tabular}                  & \begin{tabular}[c]{@{}c@{}}0.4040\\ (4)\end{tabular}           & \begin{tabular}[c]{@{}c@{}}0.4956\\ (3)\end{tabular}           & \begin{tabular}[c]{@{}c@{}}0.4221\\ (15)\end{tabular}                  & \begin{tabular}[c]{@{}c@{}}0.6409\\ (5)\end{tabular}           & \begin{tabular}[c]{@{}c@{}}0.4612\\ (3)\end{tabular}           & \begin{tabular}[c]{@{}c@{}}0.3929\\ (14)\end{tabular}         & \begin{tabular}[c]{@{}c@{}}0.5964\\ (5)\end{tabular}           \\ \hline
5             & NLP HUST                           & \begin{tabular}[c]{@{}c@{}}0.2689\\ (5)\end{tabular}           & \begin{tabular}[c]{@{}c@{}}0.2773\\ (4)\end{tabular}                   & \begin{tabular}[c]{@{}c@{}}0.2829\\ (12)\end{tabular}          & \begin{tabular}[c]{@{}c@{}}0.4732\\ (6)\end{tabular}           & \begin{tabular}[c]{@{}c@{}}0.4903\\ (5)\end{tabular}                   & \begin{tabular}[c]{@{}c@{}}0.4836\\ (12)\end{tabular}          & \begin{tabular}[c]{@{}c@{}}0.4373\\ (5)\end{tabular}           & \begin{tabular}[c]{@{}c@{}}0.4537\\ (5)\end{tabular}          & \begin{tabular}[c]{@{}c@{}}0.4465 \\ (12)\end{tabular}         \\ \hline
\textit{6}    & \textit{Extractive baseline}       & \textit{\begin{tabular}[c]{@{}c@{}}0.2625\\ (6)\end{tabular}}  & \textit{\begin{tabular}[c]{@{}c@{}}0.2464\\ (7)\end{tabular}}          & \textit{\begin{tabular}[c]{@{}c@{}}0.3174\\ (8)\end{tabular}}  & \textit{\begin{tabular}[c]{@{}c@{}}0.4772\\ (5)\end{tabular}}  & \textit{\begin{tabular}[c]{@{}c@{}}0.4582\\ (9)\end{tabular}}          & \textit{\begin{tabular}[c]{@{}c@{}}0.5391\\ (8)\end{tabular}}  & \textit{\begin{tabular}[c]{@{}c@{}}0.4339\\ (6)\end{tabular}}  & \textit{\begin{tabular}[c]{@{}c@{}}0.4164\\ (9)\end{tabular}} & \textit{\begin{tabular}[c]{@{}c@{}}0.4905\\ (9)\end{tabular}}  \\ \hline
\textit{7}    & \textit{Rule-based baseline}       & \textit{\begin{tabular}[c]{@{}c@{}}0.2611\\ (7)\end{tabular}}  & \textit{\begin{tabular}[c]{@{}c@{}}0.2634\\ (5)\end{tabular}}          & \textit{\begin{tabular}[c]{@{}c@{}}0.2947\\ (10)\end{tabular}} & \textit{\begin{tabular}[c]{@{}c@{}}0.4627\\ (8)\end{tabular}}  & \textit{\begin{tabular}[c]{@{}c@{}}0.4601\\ (8)\end{tabular}}          & \textit{\begin{tabular}[c]{@{}c@{}}0.5053\\ (11)\end{tabular}} & \textit{\begin{tabular}[c]{@{}c@{}}0.4273\\ (8)\end{tabular}}  & \textit{\begin{tabular}[c]{@{}c@{}}0.4257\\ (7)\end{tabular}} & \textit{\begin{tabular}[c]{@{}c@{}}0.4659\\ (11)\end{tabular}} \\ \hline
8             & VNU Brothers                       & \begin{tabular}[c]{@{}c@{}}0.2544\\ (8)\end{tabular}           & \begin{tabular}[c]{@{}c@{}}0.3030\\ (2)\end{tabular}                   & \begin{tabular}[c]{@{}c@{}}0.2406\\ (14)\end{tabular}          & \begin{tabular}[c]{@{}c@{}}0.4595\\ (9)\end{tabular}           & \begin{tabular}[c]{@{}c@{}}0.5315\\ (2)\end{tabular}                   & \begin{tabular}[c]{@{}c@{}}0.4312\\ (17)\end{tabular}          & \begin{tabular}[c]{@{}c@{}}0.4194\\ (12)\end{tabular}          & \begin{tabular}[c]{@{}c@{}}0.4850\\ (2)\end{tabular}          & \begin{tabular}[c]{@{}c@{}}0.3937\\ (17)\end{tabular}          \\ \hline
9             & FCoin                              & \begin{tabular}[c]{@{}c@{}}0.2544\\ (8)\end{tabular}           & \begin{tabular}[c]{@{}c@{}}0.2307\\ (9)\end{tabular}                   & \begin{tabular}[c]{@{}c@{}}0.3027\\ (9)\end{tabular}           & \begin{tabular}[c]{@{}c@{}}0.4697\\ (7)\end{tabular}           & \begin{tabular}[c]{@{}c@{}}0.4302\\ (12)\end{tabular}                  & \begin{tabular}[c]{@{}c@{}}0.5411\\ (7)\end{tabular}           & \begin{tabular}[c]{@{}c@{}}0.4296\\ (7)\end{tabular}           & \begin{tabular}[c]{@{}c@{}}0.3941\\ (13)\end{tabular}         & \begin{tabular}[c]{@{}c@{}}0.4938\\ (8)\end{tabular}           \\ \hline
10            & vts                                & \begin{tabular}[c]{@{}c@{}}0.2448\\ (9)\end{tabular}           & \begin{tabular}[c]{@{}c@{}}0.2114\\ (15)\end{tabular}                  & \begin{tabular}[c]{@{}c@{}}0.3188\\ (7)\end{tabular}           & \begin{tabular}[c]{@{}c@{}}0.4516\\ (12)\end{tabular}          & \begin{tabular}[c]{@{}c@{}}0.4048\\ (18)\end{tabular}                  & \begin{tabular}[c]{@{}c@{}}0.5438\\ (6)\end{tabular}           & \begin{tabular}[c]{@{}c@{}}0.4208\\ (10)\end{tabular}          & \begin{tabular}[c]{@{}c@{}}0.3768\\ (18)\end{tabular}         & \begin{tabular}[c]{@{}c@{}}0.5074\\ (6)\end{tabular}           \\ \hline
11            & Blue Sky                           & \begin{tabular}[c]{@{}c@{}}0.2412\\ (10)\end{tabular}          & \begin{tabular}[c]{@{}c@{}}0.2384\\ (8)\end{tabular}                   & \begin{tabular}[c]{@{}c@{}}0.2610\\ (13)\end{tabular}          & \begin{tabular}[c]{@{}c@{}}0.4588\\ (10)\end{tabular}          & \begin{tabular}[c]{@{}c@{}}0.4604\\ (7)\end{tabular}                   & \begin{tabular}[c]{@{}c@{}}0.4761\\ (13)\end{tabular}          & \begin{tabular}[c]{@{}c@{}}0.4194\\ (12)\end{tabular}          & \begin{tabular}[c]{@{}c@{}}0.4205\\ (8)\end{tabular}          & \begin{tabular}[c]{@{}c@{}}0.4358\\ (13)\end{tabular}          \\ \hline
12            & HUSTLANG                           & \begin{tabular}[c]{@{}c@{}}0.2361\\ (11)\end{tabular}          & \begin{tabular}[c]{@{}c@{}}0.2880\\ (3)\end{tabular}                   & \begin{tabular}[c]{@{}c@{}}0.2157\\ (17)\end{tabular}          & \begin{tabular}[c]{@{}c@{}}0.4360\\ (16)\end{tabular}          & \begin{tabular}[c]{@{}c@{}}0.5176\\ (4)\end{tabular}                   & \begin{tabular}[c]{@{}c@{}}0.3981\\ (18)\end{tabular}          & \begin{tabular}[c]{@{}c@{}}0.4000\\ (15)\end{tabular}          & \begin{tabular}[c]{@{}c@{}}0.4750\\ (3)\end{tabular}          & \begin{tabular}[c]{@{}c@{}}0.3651\\ (18)\end{tabular}          \\ \hline
13            & SGSUM                              & \begin{tabular}[c]{@{}c@{}}0.2322\\ (12)\end{tabular}          & \begin{tabular}[c]{@{}c@{}}0.2106\\ (16)\end{tabular}                  & \begin{tabular}[c]{@{}c@{}}0.2896\\ (11)\end{tabular}          & \begin{tabular}[c]{@{}c@{}}0.4575\\ (11)\end{tabular}          & \begin{tabular}[c]{@{}c@{}}0.4279\\ (13)\end{tabular}                  & \begin{tabular}[c]{@{}c@{}}0.5282\\ (10)\end{tabular}          & \begin{tabular}[c]{@{}c@{}}0.4235\\ (9)\end{tabular}           & \begin{tabular}[c]{@{}c@{}}0.3954\\ (12)\end{tabular}         & \begin{tabular}[c]{@{}c@{}}0.4897\\ (10)\end{tabular}          \\ \hline
14            & vc-datamining                      & \begin{tabular}[c]{@{}c@{}}0.2304\\ (13)\end{tabular}          & \begin{tabular}[c]{@{}c@{}}0.1663\\ (20)\end{tabular}                  & \begin{tabular}[c]{@{}c@{}}0.4371\\ (3)\end{tabular}           & \begin{tabular}[c]{@{}c@{}}0.4496\\ (14)\end{tabular}          & \begin{tabular}[c]{@{}c@{}}0.3450\\ (20)\end{tabular}                  & \begin{tabular}[c]{@{}c@{}}0.7036\\ (2)\end{tabular}           & \begin{tabular}[c]{@{}c@{}}0.4201\\ (11)\end{tabular}          & \begin{tabular}[c]{@{}c@{}}0.3218\\ (20)\end{tabular}         & \begin{tabular}[c]{@{}c@{}}0.6590\\ (2)\end{tabular}           \\ \hline
15            & TCV-AI                             & \begin{tabular}[c]{@{}c@{}}0.2288\\ (14)\end{tabular}          & \begin{tabular}[c]{@{}c@{}}0.1687\\ (19)\end{tabular}                  & \begin{tabular}[c]{@{}c@{}}0.3976\\ (5)\end{tabular}           & \begin{tabular}[c]{@{}c@{}}0.4502\\ (13)\end{tabular}          & \begin{tabular}[c]{@{}c@{}}0.3485\\ (19)\end{tabular}                  & \begin{tabular}[c]{@{}c@{}}0.6813\\ (3)\end{tabular}           & \begin{tabular}[c]{@{}c@{}}0.4190\\ (13)\end{tabular}          & \begin{tabular}[c]{@{}c@{}}0.3245\\ (19)\end{tabular}         & \begin{tabular}[c]{@{}c@{}}0.6340 \\ (3)\end{tabular}          \\ \hline
16            & Team Attention                     & \begin{tabular}[c]{@{}c@{}}0.2131\\ (15)\end{tabular}          & \begin{tabular}[c]{@{}c@{}}0.2159\\ (14)\end{tabular}                  & \begin{tabular}[c]{@{}c@{}}0.2265\\ (16)\end{tabular}          & \begin{tabular}[c]{@{}c@{}}0.4274\\ (18)\end{tabular}          & \begin{tabular}[c]{@{}c@{}}0.4251\\ (14)\end{tabular}                  & \begin{tabular}[c]{@{}c@{}}0.4514\\ (15)\end{tabular}          & \begin{tabular}[c]{@{}c@{}}0.3848\\ (18)\end{tabular}          & \begin{tabular}[c]{@{}c@{}}0.3835\\ (17)\end{tabular}         & \begin{tabular}[c]{@{}c@{}}0.4056\\ (15)\end{tabular}          \\ \hline
17            & Cyber Intellect                    & \begin{tabular}[c]{@{}c@{}}0.2116\\ (16)\end{tabular}          & \begin{tabular}[c]{@{}c@{}}0.2085\\ (17)\end{tabular}                  & \begin{tabular}[c]{@{}c@{}}0.2270\\ (15)\end{tabular}          & \begin{tabular}[c]{@{}c@{}}0.4464\\ (15)\end{tabular}          & \begin{tabular}[c]{@{}c@{}}0.4468\\ (10)\end{tabular}                  & \begin{tabular}[c]{@{}c@{}}0.4627\\ (14)\end{tabular}          & \begin{tabular}[c]{@{}c@{}}0.4028\\ (14)\end{tabular}          & \begin{tabular}[c]{@{}c@{}}0.4030\\ (10)\end{tabular}         & \begin{tabular}[c]{@{}c@{}}0.4177\\ (14)\end{tabular}          \\ \hline
18            & HHH                                & \begin{tabular}[c]{@{}c@{}}0.1919\\ (17)\end{tabular}          & \begin{tabular}[c]{@{}c@{}}0.1915\\ (18)\end{tabular}                  & \begin{tabular}[c]{@{}c@{}}0.2076\\ (18)\end{tabular}          & \begin{tabular}[c]{@{}c@{}}0.4228\\ (19)\end{tabular}          & \begin{tabular}[c]{@{}c@{}}0.4350\\ (11)\end{tabular}                  & \begin{tabular}[c]{@{}c@{}}0.4336\\ (16)\end{tabular}          & \begin{tabular}[c]{@{}c@{}}0.3888\\ (16)\end{tabular}          & \begin{tabular}[c]{@{}c@{}}0.4005\\ (11)\end{tabular}         & \begin{tabular}[c]{@{}c@{}}0.3984\\ (16)\end{tabular}          \\ \hline
\textit{19}   & \textit{Anchor baseline}           & \textit{\begin{tabular}[c]{@{}c@{}}0.1886\\ (18)\end{tabular}} & \textit{\begin{tabular}[c]{@{}c@{}}0.2306\\ (10)\end{tabular}}         & \textit{\begin{tabular}[c]{@{}c@{}}0.1734\\ (19)\end{tabular}} & \textit{\begin{tabular}[c]{@{}c@{}}0.4321\\ (17)\end{tabular}} & \textit{\begin{tabular}[c]{@{}c@{}}0.5210\\ (3)\end{tabular}}          & \textit{\begin{tabular}[c]{@{}c@{}}0.3900\\ (19)\end{tabular}} & \textit{\begin{tabular}[c]{@{}c@{}}0.3869\\ (17)\end{tabular}} & \textit{\begin{tabular}[c]{@{}c@{}}0.4659\\ (4)\end{tabular}} & \textit{\begin{tabular}[c]{@{}c@{}}0.3498\\ (19)\end{tabular}} \\ \hline
\textit{20}   & \textit{Abstractive baseline}      & \textit{\begin{tabular}[c]{@{}c@{}}0.1497\\ (19)\end{tabular}} & \textit{\textbf{\begin{tabular}[c]{@{}c@{}}0.3061\\ (1)\end{tabular}}} & \textit{\begin{tabular}[c]{@{}c@{}}0.1025\\ (20)\end{tabular}} & \textit{\begin{tabular}[c]{@{}c@{}}0.3226\\ (20)\end{tabular}} & \textit{\textbf{\begin{tabular}[c]{@{}c@{}}0.5801\\ (1)\end{tabular}}} & \textit{\begin{tabular}[c]{@{}c@{}}0.2299\\ (20)\end{tabular}} & \textit{\begin{tabular}[c]{@{}c@{}}0.2895\\ (19)\end{tabular}} & \textit{\begin{tabular}[c]{@{}c@{}}\textbf{0.5205}\\ (1)\end{tabular}} & \textit{\begin{tabular}[c]{@{}c@{}}0.2065\\ (20)\end{tabular}} \\ \hline
\end{tabular}

}
\caption{The official results on the Private Test. The number highlighted in bold is the highest result in each column. The number in the bracket () is the corresponding rank of a score. Baseline results are shown in italic.}
\end{table*}

\end{document}